%% file: acl_latex.tex
\pdfoutput=1

\documentclass[11pt]{article}

\usepackage[final]{acl}

\usepackage{times}
\usepackage{xcolor}
\usepackage{xspace}
\usepackage[frozencache,cachedir=.]{minted}

\usepackage{latexsym}
\usepackage{bbm}
\usepackage{makecell}
\definecolor{cexample}{rgb}{0.23, 0.30, 0.45}
\definecolor{ctemplate}{rgb}{0.23, 0.45, 0.30}

\newcommand{\ra}[1]{\renewcommand{\arraystretch}{#1}}
\usepackage[T1]{fontenc}

\usepackage[utf8]{inputenc}

\usepackage{microtype}

\usepackage{url}
\usepackage{framed}
\usepackage{multirow}
\usepackage{graphicx}
\usepackage{caption}
\usepackage{subcaption}
\usepackage{dsfont}
\usepackage{centernot}
\usepackage{amsmath}
\usepackage{todonotes}
\usepackage[normalem]{ulem}

\usepackage[utf8]{inputenc} 
\usepackage[T1]{fontenc}    
\usepackage{hyperref}       
\usepackage{url}            
\usepackage{booktabs}       
\usepackage{amsfonts}       
\usepackage{nicefrac}       
\usepackage{microtype}      

\usepackage{float}
\usepackage{siunitx} 
\usepackage{etoolbox} 
\usepackage{listings}

\restylefloat{table}
\renewcommand{\arraystretch}{1.2} 

\input{macros}

\definecolor{posc}{rgb}{0.90, 0.98, 0.96}
\definecolor{negc}{rgb}{1, 0.94, 0.92}

\newcommand{\reducedstrut}{\vrule width 0pt height .9\ht\strutbox depth .9\dp\strutbox\relax}

\newcommand{\colbox}[2]{  \begingroup
  \setlength{\fboxsep}{0pt}%
  \colorbox{#1}{\reducedstrut#2\/}%
  \endgroup}

\newcommand{\posb}[1]{\colbox{posc}{#1\xspace}}
\newcommand{\negb}[1]{\colbox{negc}{#1\xspace}}

\newcommand{\template}[1]{{\cmss{#1}\xspace}}
\definecolor{lightgrey}{rgb}{0.875, 0.875, 0.875}
\definecolor{lightgrey}{rgb}{0.875, 0.875, 0.875}
\definecolor{hcolor}{rgb}{0.972, 0.953, 0.823}

\newcommand{\mybox}[1]{\colbox{lightgrey}{#1}}
\newcommand{\highlight}[1]{\colbox{hcolor}{#1}}

\DeclareMathOperator*{\argmax}{arg\,max}

\title{Fixing Model Bugs with Natural Language Patches}

\author{
\textbf{Shikhar Murty}\textsuperscript{$\dagger\star$}
\hspace{.1cm}\textbf{Christopher D. Manning}\textsuperscript{$\dagger$} \hspace{.1cm} \textbf{Scott Lundberg}\textsuperscript{$\ddagger$} \hspace{.1cm} \textbf{Marco Tulio Ribeiro}\textsuperscript{$\ddagger$} \\
\textsuperscript{$\dagger$}Computer Science Department, Stanford University\quad
  \textsuperscript{$\ddagger$}Microsoft Research\\
  \texttt{\{smurty,manning\}@cs.stanford.edu, \{scott.lundberg, marcotcr\}@microsoft.com} 
}

\begin{document}
\maketitle
\renewcommand\thefootnote{}\footnote{\textsuperscript{$\star$} Part of the work done at Microsoft Research.}

\renewcommand*{\thefootnote}{\arabic{footnote}}
\setcounter{footnote}{0}

\begin{abstract}    

Current approaches for fixing systematic problems in NLP models (\egg{} regex patches, finetuning on more data) are either brittle, or labor-intensive and liable to shortcuts. In contrast, humans often provide corrections to each other through natural language. Taking inspiration from this, we explore \emph{natural language patches}---declarative statements that allow developers to provide corrective feedback at the right level of abstraction, either overriding the model (``if a review gives 2 stars, the sentiment is negative'') or providing additional information the model may lack (``if something is described as the bomb, then it is good'').
We model the task of determining if a patch applies separately from the task of integrating patch information, and show that with a small amount of synthetic data, we can teach models to effectively use real patches on real data---1 to 7 patches improve accuracy by {\textasciitilde}1--4 accuracy points on different slices of a sentiment analysis dataset, and F1 by 7 points on a relation extraction dataset.
Finally, we show that finetuning on as many as 100 labeled examples may be needed to match the performance of a small set of language patches. 

\end{abstract}

\section{Introduction}

\begin{figure}[t]
\centering
    \begin{subfigure}{.48\textwidth}
  \includegraphics[width=0.99\linewidth]{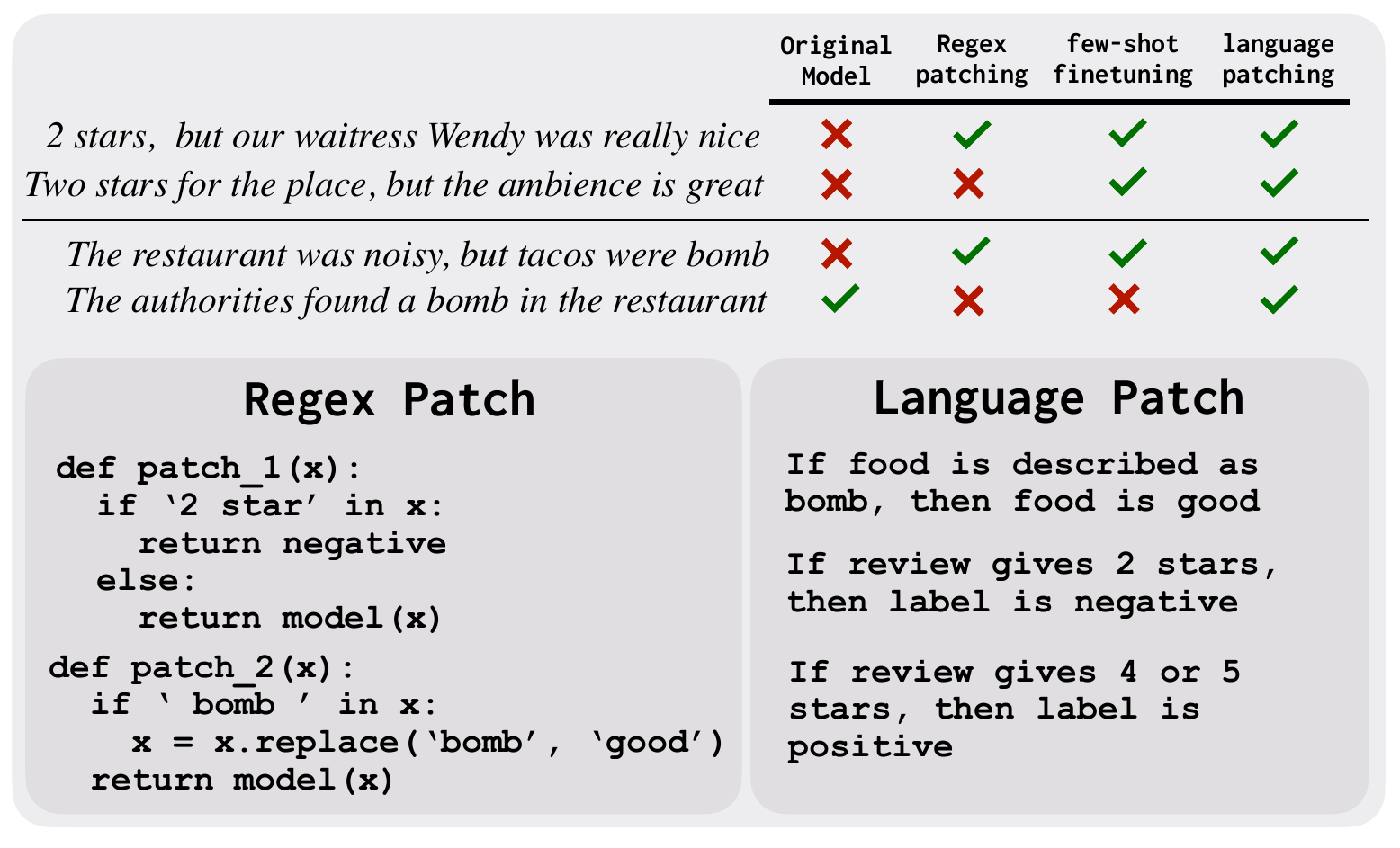}
  \caption{Developer identifies bugs in model}
  \label{overview:a}
\end{subfigure}%
\vspace{1em}
\begin{subfigure}{.48\textwidth}
  \includegraphics[width=.99\linewidth]{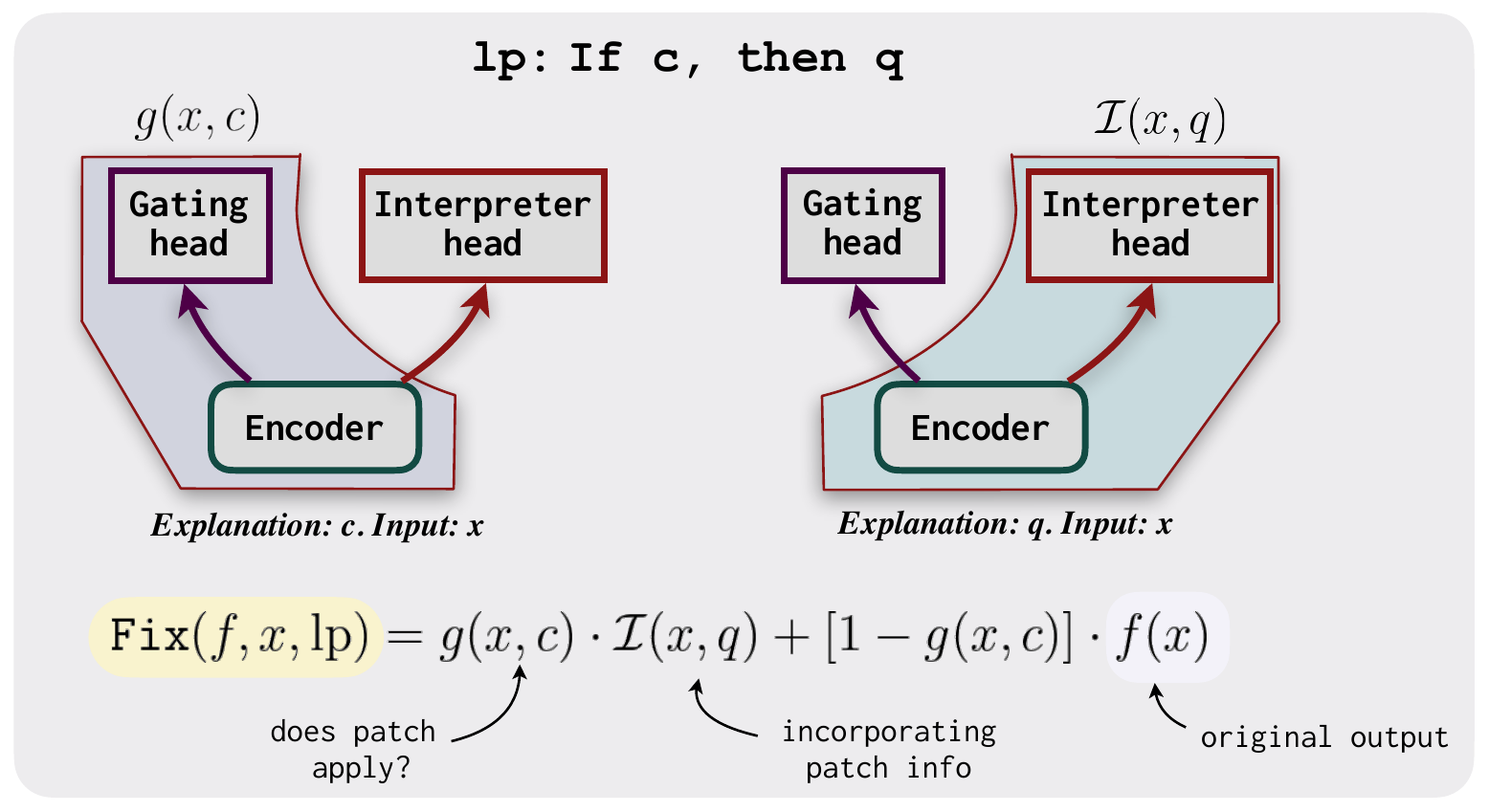}
  \caption{Incorporating Language Patches}
  \label{overview:b}
\end{subfigure}
\caption{(\subref{overview:a}) Developers typically fix bugs by writing brittle regex patches or by finetuning on additional data, which is prone to simple shortcuts. In contrast, natural language patches are more expressive than regexes and prevent shortcuts by abstractly specifying when they should be applied. (\subref{overview:b}) Our proposed model uses a gating head to predict whether a patch condition $\cond$ applies to the input. That (soft) prediction is then used to combine the original model output with the output of an interpreter head that uses textual features from both the input as well as the patch consequent $\cons$.}
\label{fig:overview}
\end{figure}

Natural language enables humans to communicate a lot at once with shared abstractions.
For example, in teaching someone about the colloquial use of the term ``bomb'', we might say \textit{describing food as `bomb' means it is very good, while saying someone bombed means it was disappointing}.
This simple sentence uses various abstractions (\egg{} ``food'') to provide context-dependent information, making it easy for humans to generalize and understand sentences such as ``The tacos were bomb'' or ``The chef bombed'' without ever having seen such examples.

There is a growing body of research focused on using language to give instructions, supervision and even inductive biases to models instead of relying exclusively on labeled examples, \egg{} building neural representations from language descriptions \citep{Andreas2018, Murty2020, Mu2020}, or language / prompt-based zero-shot learning  \citep{brown2020language, hanjie2022semantic, chen21}. However, language is yet to be successfully applied for \emph{corrective} purposes, where the user interacts with an existing model to improve it. As shown in \figref{overview:a}, if a developer discovers that a model contains bugs \cite[\iee{} systematic errors;][]{ribeiro-etal-2020-beyond}, common fixes are either brittle regex-based patches (\egg{}  \figref{overview:a} left, where patches either override predictions or replace the word ``bomb'' with the word ``good''), or collecting hundreds of additional datapoints for finetuning, a tedious and computationally demanding process that can still lead to shortcuts such as assuming the word ``bomb'' is always positive (\egg{} if the additional finetuning data mostly has the word in its colloquial sense). Instead, we envision a setting where developers provide corrective feedback through a \emph{Natural Language Patch}---a concise statement such as \explanation{If food is described as bomb, then food is good}. Language makes it easy for developers to express feedback at the right level of abstraction without having to specify exactly how the condition is applied. The patching system is responsible for applying the patch and integrating the information appropriately, \egg{} applying it to  ``The tacos were the bomb'' but not to ``The authorities found a bomb in the restaurant''.

In this work, we present an approach for patching neural models with natural language. Any patching system has to determine when a patch is relevant, and how it should modify model behavior.
We model these tasks separately (\figref{overview:b}): a \emph{gating} head soft-predicts whether the patch should be applied (\egg{} ``food is described as bomb''), and an \emph{interpreter} head predicts a new output by combining the information in the patch (\egg{} ``food is good'') with the original input. Both heads are trained on \emph{synthetic data} in a \emph{patch tuning} stage between training and deployment, such that new patches can be combined into a library of patches (or maybe various user-specific libraries), and applied at test-time without further training.
In addition to the expressivity provided by abstractions, language-based patching is \emph{lightweight}, \emph{iterative} and easily \emph{reversible}. Much like software, developers can write / edit / remove patches iteratively until errors on unit tests or validation data are fixed, without constantly retraining the model.

Our experiments are organized as follows. First, in Section~\ref{sec:synth_experiments}, we present controlled experiments that indicate these patches work even for \emph{abstract} conditions, where regex patches would be infeasible or very difficult---that is, they are applied correctly when the patch condition is met, and do nothing otherwise. Perhaps surprisingly, this is true even for test-time patches that are very different than the ones used in the patch finetuning stage. Next, in Section~\ref{sec:real_experiments}, we show that despite the synthetic nature of the patch tuning phase, a small set of very simple patches can fix bugs (and thus improve performance) on real benchmarks for two different tasks---1 to 6 simple language patches improve performance by {\textasciitilde}1--4 accuracy points on two slices from the Yelp reviews dataset, while 7 patches improve performance by \textasciitilde{}7 F1 points on a relation extraction task derived from NYT. Finally, in Section~\ref{sec:comparison_with_finetuning}, we compare language patching, a computationally \emph{lightweight} procedure, with finetuning, a computationally and human-labor \emph{intensive} procedure, and find that as many as 100 labeled examples are needed to match performance gains from a small set of 1 to 7 patches. Further, finetuning sometimes fixes bugs at the expense of introducing new bugs, while patches maintain prior performance on inputs where they do not apply. 

\section{Natural Language Patching}
\label{sec:our_approach}

\paragraph{Setup.} We are given a model $\model$, mapping an input text $x$ to a probability distribution over its output space, $\model(x) = \Pr(y \mid x)$. The model contains \emph{bugs}---defined as behaviors inconsistent with users' preferences or the ``ground truth''--- which we want to fix with a library of patches $\patchlib = \{\lpp{1}, \lpp{2}, \ldots, \lpp{t}\}$.
Users explicitly indicate the condition under which each patch applies and the consequence of applying it, such that each patch is in the form \explanation{If (condition) \cond{}, then (consequence) \cons{}}. We use this format to make modeling easier, noting that it still allows for very flexible patching through high level abstractions (\egg{} ``if the customer complains about the ambience'', ``if food is not mentioned'', etc), and that most patches have an implicit applicability function, and thus can be converted to this format.

\begin{table*}
\centering
\ra{1.3}
\scriptsize
\begin{tabular}{@{}cm{4cm}l@{}}
\toprule
                                          & \textbf{Template} & \textbf{Examples} \\ \midrule
                  
\multirow{6}{*}{\textit{Patches}}         &  \textbf{\em Override:} \template{If \mybox{aspect} is good, then label is positive}       &  \makecell[ml]{\eexample{$e_0$}{If service is good, then label is positive} \\\eexample{$e_1$}{If food is good, then label is positive}{}}        \\ \addlinespace

                                          & \textbf{\em Override:}  \template{If \mybox{aspect} is bad, then label is negative}         &    \makecell[ml]{\eexample{$e_2$}{If service is bad, then label is negative} \\\eexample{$e_3$}{If ambience is bad then label is negative}}     \\ \addlinespace
                                          & \textbf{\em Override:} \template{If review contains words like \mybox{word}, then label is positive}      & \makecell[ml]{\eexample{$e_4$}{If review contains words like zubin, then label is positive} \\\eexample{$e_5$}{If review contains words like excellent, then label is positive}}         \\ \addlinespace
                                          & \textbf{\em Override:} \template{If review contains words like \mybox{word}, then label is negative}         &  \makecell[ml]{\eexample{$e_6$}{If review contains words like wug, then label is negative} \\\eexample{$e_7$}{If review contains words like really bad, then label is negative}}         \\  \addlinespace
    & \textbf{\em Feature Based:} \template{If \mybox{aspect} is described as \mybox{word}, then \mybox{aspect} is good / bad} &  \makecell[ml]{\eexample{$e_8$}{If food is described as above average, then food is good} \\\eexample{$e_9$}{If food is described as wug, then food is bad} \\\eexample{$e_{10}$}{If food is described as zubin, then service is good} \\\eexample{$e_{11}$}{If service is described as not great, then service is bad}}         \\ \midrule
\multirow{6}{*}{\textit{Inputs}} &    \template{The \mybox{aspect} at the restaurant was \mybox{adj}}  &  \makecell[ml]{\inp{The service at the restaurant was really good.}{\posb{$e_0$}, \negb{$e_3$}} \\\inp{The food at the restaurant was wug.}{\posb{$e_6$}, \posb{$e_9$}} }         \\ \addlinespace

                    &  \template{The  restaurant had \mybox{adj} \mybox{aspect}}        &  \makecell[ml]{\inp{The restaurant had really bad service.}{\posb{$e_7$}, \posb{$e_2$}, \negb{$e_{11}$}} \\\inp{The restaurant had zubin ambience.}{\posb{$e_4$}, \negb{$e_{10}$}}}        \\ \addlinespace

                    &  \template{The \mybox{aspect1} was \mybox{adj1}, the \mybox{aspect2} was \mybox{adj2}}        &  \makecell[ml]{\inp{The food was good, the ambience was bad.}{\posb{$e_1$}, \posb{$e_3$}, \negb{$e_1$}} \\\inp{The service was good, the food was not good.}{\posb{$e_0$}, \negb{$e_1$}}}        \\ \addlinespace

                    &  \template{The \mybox{aspect1} was \mybox{adj1} but the \mybox{aspect2} was really \mybox{adj2}}        &  \makecell[ml]{\inp{The food was good, but the service was really bad.}{\posb{$e_7$}, \posb{$e_1$}, \negb{$e_0$}} \\\inp{The ambience was bad, but the food was really not wug.}{\posb{$e_3$}, \negb{$e_9$}}}         \\  \addlinespace

                    &  \template{The \mybox{aspect1} was really \mybox{adj1} even though \mybox{aspect2} was \mybox{adj2}}       &   \makecell[ml]{\inp{The food was really bad even though the ambience was excellent.}{\posb{$e_5$}, \posb{$e_7$}, \negb{$e_8$}} \\\inp{The food was really zubin, even though the service was bad}{\posb{$e_{4}$}}, \posb{$e_{10}$}, \negb{$e_0$}}       \\ \bottomrule
\end{tabular}
\caption{Patch and Input templates used for the Patch Finetuning stage for the sentiment analysis task. We divide our patches into 2 categories: \emph{Override} and \emph{Feature Based} (see Section~\ref{sec:our_approach} for more details). For each input, we provide examples of patches that \posb{apply} and patches that \negb{don't apply}. The simplistic nature of these templates makes them easy to write without access to additional data sources or lexicons.}
\label{tab:synthsent}
\end{table*}

\paragraph{Applying Patches.}
As indicated in \figref{overview:b}, our model consists of two separate heads. The gating head $\rmodel$ computes the probability that the condition specified by $\lp = (\cond, \cons)$ is true for a given input $x$ as $g(x, c)$. The interpreter head $\imodel$ computes a new distribution over the label space, that conditions on $x$ \emph{and} the consequence $\cons$. This is then combined with the original model output $\model(x)$ using the above gating probability.
A single patch $\lp = (\cond, \cons)$, can be applied to any input $x$ as
\begin{align}
    \text{\fixed{\lp}} &= \rmodel(x, c) \cdot \imodel(x, q) \label{eq:patching} \\ \notag &+ [1 - \rmodel(x,c)] \cdot \model(x). 
\end{align}

Given a library of patches $\patchlib = \{\lpp{1}, \ldots, \lpp{t}\}$, we find the most relevant patch $\lp^{*}$ for the given input, and use that to update the model, 
\begin{align}
    \lp^{*} &= \argmax_{\lpp{i} \in \patchlib} g(x, c_i), \label{eq:scalability} \\
    \text{\fixed{\patchlib}} &= \text{\fixed{$\lp^{*}$}}.
\end{align}

\paragraph{Patch Types.} We consider two categories of patches (examples in Table~\ref{tab:synthsent}). \textbf{Override} patches are of the form \explanation{If \mybox{cond}, then label is $l$} \iee{} they override the model's prediction on an input if the patch condition is true. For these patches, we do not use the interpreter head since $\imodel(x, \text{``label is $l$''}) = l$. \textbf{Feature-based} patches are of the form \explanation{If \mybox{cond}, then \mybox{feature}}, \iee{} they provide the model with a contextual feature ``hint'' in natural language, \egg{} in \figref{fig:synth_data} the \mybox{feature} is ``food is good''. For these patches, the model needs to integrate the hints with the original data, and thus both the gating and interpreter heads are used.

\section{Training Patchable Models}
\label{sec:our_approach2}
Assuming $\model$ has a text encoder and a classification head, we have two finetuning stages. In the \textbf{Task Finetuning} stage, we train $\model$ on a labeled dataset $\{x_i, y_i\}$ (standard supervised learning). In the \textbf{Patch Finetuning} stage, we use the learnt encoder and learn $\rmodel$ (initialized randomly) and $\imodel$ (initialized with the classification head). For the patch finetuning stage, we write a small set of patch templates covering the kinds of patches users may write for their own application (see Table~\ref{tab:synthsent} for the patch templates used for our sentiment analysis results). Based on these templates, we instantiate a small number of patches along with synthetic labeled examples. This gives us a dataset $\{x_i, y_i, \lp_i\}$, where $\lpp{i}$ consists of a condition $\condp{i}$  as well as a consequence $\cons_i$. The interpreter head $\imodel$ is trained to model $\Pr(y_i \mid x_i, \cons_i)$ through standard log-likelihood maximization. The gating head $\rmodel$ is trained via noise contrastive estimation to maximize
\begin{align}
\tiny
\log \rmodel(x_i, \condp{i}) - \sum_{\condp{j} \in \negex{x_i}} \log \rmodel(x_i, \condp{j}),
\end{align}
where $\negex{x_i}$ is a randomly sampled set of negative conditions for $x_i$.

\begin{figure}[]
    \begin{subfigure}{.24\textwidth}
  \includegraphics[width=0.99\linewidth]{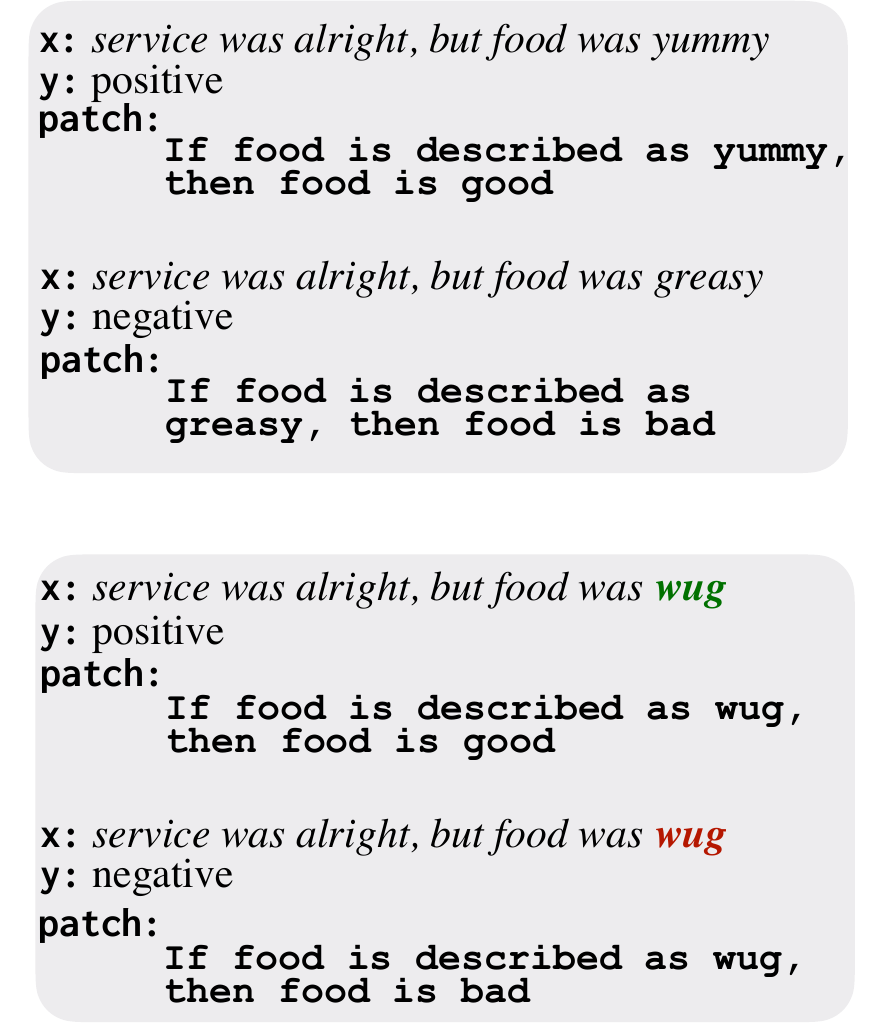}
  \caption{}
  \label{fig:a}
\end{subfigure}%
\begin{subfigure}{.24\textwidth}
  \includegraphics[width=.99\linewidth]{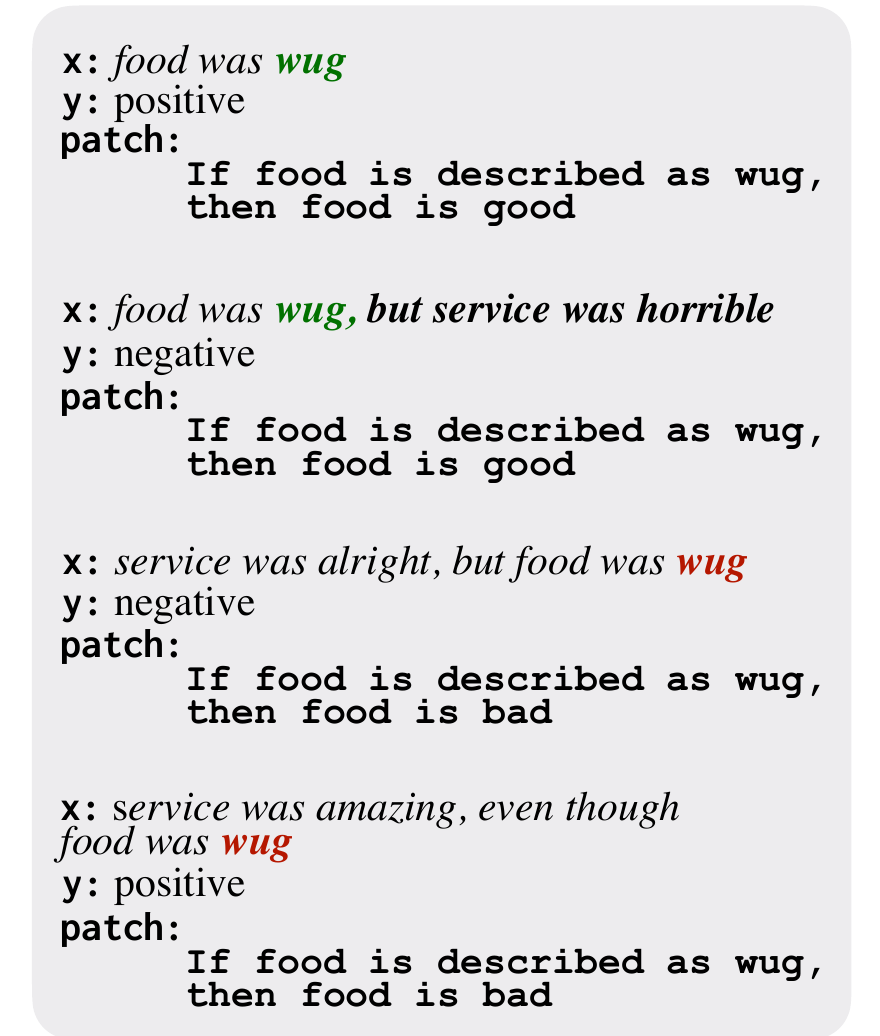}
  \caption{}
  \label{fig:b}
\end{subfigure}
\caption{A model can learn from just the labels that ``yummy'' and ``greasy'' are positive and negative words respectively, and learn to perfectly fit training data without ever using patch features (\subref{fig:a}, top). This behavior can be explicitly prevented via EITs (\subref{fig:a}, bottom). A model may also fit the data without using the input features by always predicting 1 / 0 for ``food is good'' / ``food is bad'' (\subref{fig:a}, top/bottom). Thus, we additionally ensure that the label cannot be inferred from the patch alone (\subref{fig:b}).} 
\end{figure}

\paragraph{Entropy Increasing Transformations.} Patch Finetuning will fail if the synthetic data can be fit by a model that ignores the input or the patch (\figref{fig:a}). Thus, to ensure our model cannot fit the synthetic data without combining patch features with inputs, we perturb the inputs with \emph{Entropy Increasing Transformations} (EITs). We identify words from the input template for which the patch supplies additional information \egg{} aspect adjectives, relationship between entities, and transform these into a small set of \emph{nonce} words. Crucially, the meanings of these nonce words vary  from example to example, and can only be inferred from the patch (\figref{fig:a} bottom; more examples in Appendix~\ref{sec:appendix_re}). Intuitively, the transformations inject an additional source of randomness which can only be recovered via the patch features. Such transformations are also used in \citet{rajendran20} in the context of meta-learning. EITs alone do not fix the failure mode where the model can fit the data without using input features at all. For example, in \figref{fig:a} bottom, the model might learn a shortcut so that it always predicts 1/0 for ``food is good'' / ``food is bad'', regardless of the input. Thus, in addition to EITs, to ensure that the model uses input features, we ensure that a given patch consequence $q$ and the target label are independent (\figref{fig:b}).

\section{Experimental Setup}
\label{sec:experiments}

\paragraph{Applications.} We apply our method to binary sentiment analysis and relation extraction. For sentiment analysis, our task finetuning data comes from SST2 \cite{socher-etal-2013-recursive}. For relation extraction, we use the {\spouse} dataset \citep{Hancock2018} for task finetuning, where the objective is to determine whether two entities are married or not given a textual context about them.

\paragraph{Model.}  We use T5-large \citep{raffel2019exploring} as implemented in the transformers library \citep{wolf-etal-2020-transformers} for all  experiments. Both the gating and interpreter heads are separate decoders learnt on top of a shared encoder and each of these components are initialized with the corresponding T5 pre-trained weights. To prevent catastrophic forgetting on the original task during patch finetuning, we also multi-task learn the patch finetuning loss along with the original task loss. Templates for generating patches for patch finetuning are in Table~\ref{tab:synthsent} for sentiment analysis and in Table~\ref{tab:synthspouse} ( Section~\ref{sec:appendix_re}) for relation extraction. We train separate models for override and feature-based patches (the former does not need an interpreter head). When using a patch, its content (either \cond{} for the gating head or \cons{} for the interpreter head) is inserted in the beginning of the input with a separator as in \figref{overview:b}.

\paragraph{Baselines.}  We report performance of the original model with only task finetuning (\original) and the model obtained after patch finetuning (\originalpf) without using any patches, to isolate the gains of language patches from those induced by training on additional synthetic data. We also report results obtained from prompting \original{} with our patches (\originalprompts), \iee{} inserting the patch text before the input text to see how well finetuned T5 follows instructions. To use multiple patches for this baseline, we prompt the model with each individual patch and ensemble results with majority voting. Finally, we experiment with \emph{regex-based} patches (\rbpatches{}) where patch conditions are converted to regex rules and consequents are converted into functions $\rulef_\cons(x)$. For override patches, this function simply outputs the specified label. For sentiment analysis, where feature based patches supply contextual meanings, $\rulef_\cons(x)$ replaces words with specified meanings \egg{} replacing ``bomb'' with ``good'' in ``the food was bomb''. For feature based patches on relation extraction, $\rulef_\cons(x)$ appends the patch consequent to the input text.

\section{Controlled Experiments}
\label{sec:synth_experiments}

\begin{figure}
\centering
\includegraphics[width=\linewidth]{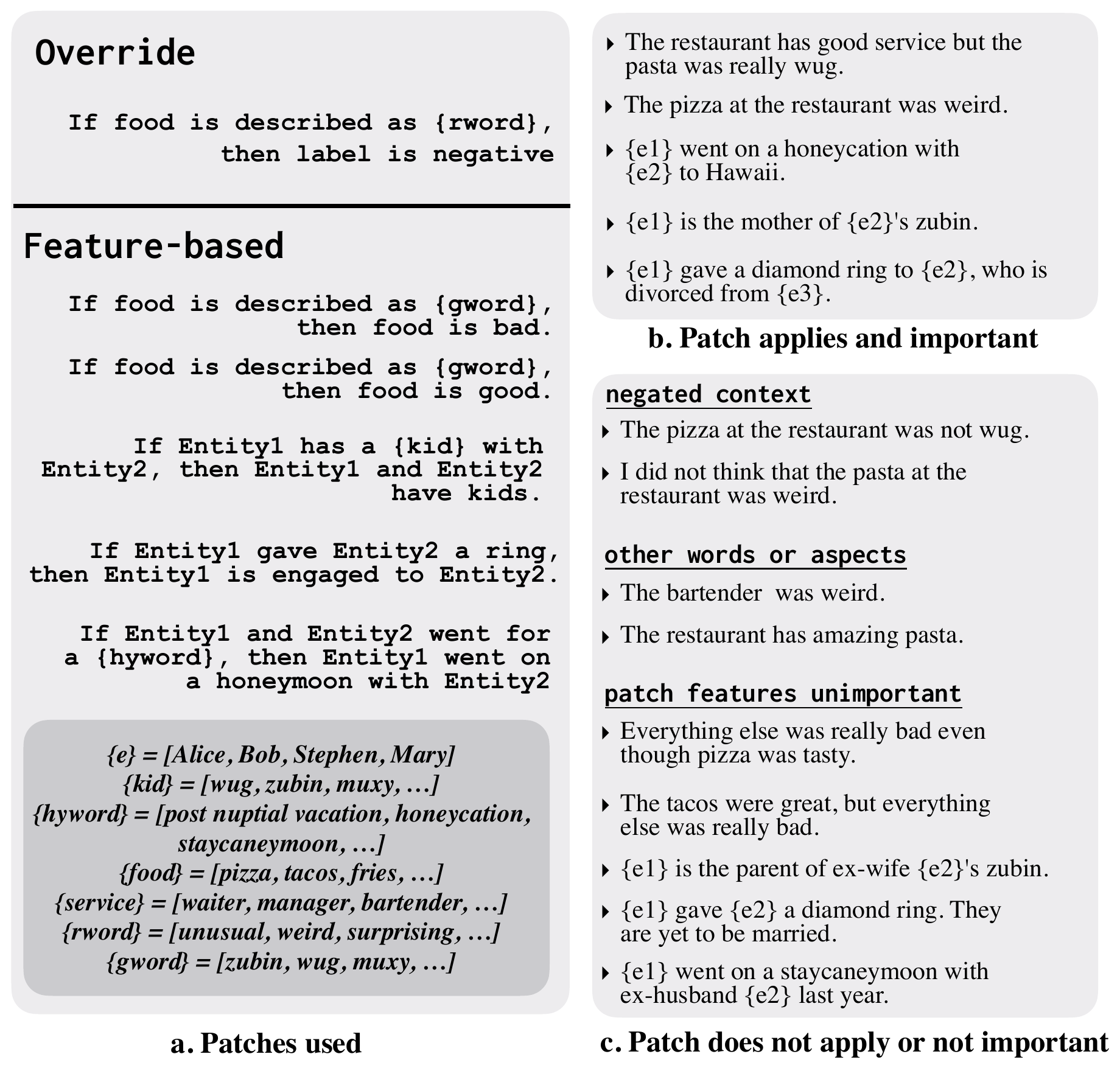}
\caption{(a) Example patches used for our controlled experiments. (b) Some inputs where the patch is important for making correct predictions. (c) To control for spurious behaviors such as copying label words from the patch, performing simple string lookups or affecting predictions when patch features are unimportant, we also construct invariance tests where we expect model predictions to be unaffected by the patch.}
\label{fig:synth_data}
\end{figure}

We test the behavior of language patches (and baselines) under different controlled conditions with CheckList \cite{ribeiro-etal-2020-beyond}. Patches and example inputs are presented in \figref{fig:synth_data}.
We test cases where patches apply \emph{and are relevant} for predictions, and corresponding cases where they either do not apply or are not relevant.
Thus, models that rely on shortcuts such as copying the label word from the patch or merely performing token matching perform poorly on the CheckList.

For sentiment analysis, we test Override patches with \emph{abstract} conditions (\egg{} ``If \emph{food} is described as weird, then label is negative'' ) on various concrete instantiations such as ``The \emph{pizza} at the restaurant was weird''. We also construct invariance tests ({\overrideabsneg}), where adding such patches should not change predictions on inputs where the condition is false (\egg{} ``The waiter was weird'', ``The tacos were not weird'').
We also construct tests for feature-based patches ({\interpretabs{}}) where patches provide meaning for nonce adjectives, with analogous invariance tests ({\interpretabsn{}}).
Finally, we construct analogous tests for relation extraction, where patches fill in \emph{reasoning gaps} in the model such as \explanation{If Entity1 gave Entity2 a ring, then Entity1 and Entity2 are engaged}.

We present the results in Table \ref{tab:results-synth}, where we first note that \originalpf{} does not perform well overall, and thus patching improvements are not merely a result of the additional synthetic data. \rbpatches{} cannot handle abstract conditions, and thus (as expected) does not change predictions on sentiment analysis, and does not do well on relation extraction. 
While merely inserting the patch into the input (\originalprompts{}) results in some gains when the patch applies, it does so at the cost of changing predictions when the patch does not apply ({\overrideabsneg{}} and {\interpretabsn{}}).
In contrast to baselines, our method is able to apply \emph{abstract} patches correctly on concrete instantiations, disregarding them when they do not apply, without relying on shortcuts such as copying the label from the consequent or merely checking for matching words between patch and input (all of which are tested by the invariance tests).

\begin{table}
\centering
\ra{1.3}
\tiny
\begin{tabular}{@{}lrrrrp{0.001cm}rr@{}} \toprule
\multirow{2}{*}{Model} & \multicolumn{4}{c}{Sentiment Analysis} && \multicolumn{2}{c@{}}{Relation Extraction} \\ 

\cmidrule{2-5} \cmidrule{7-8}  
 & \overrideabs & \overrideabsneg{} & \interpretabs{} &  \interpretabsn{} && \spousesynth{}  & \spousesynthc{} \\ \midrule

\original{} & 50.0 & n/a & 59.1 & n/a && 14.5 & n/a \\
\originalpf{} & 50.0 & n/a & 59.9 & n/a && 35.8 & n/a \\
\rbpatches{} & 50.0 & \textbf{100.0} & 59.9 & \textbf{100.0} && 45.8 & 88.1 \\
\originalprompts{} & 68.7 & 63.8 & 64.3 & 85.4 && 13.9 & 87.6 \\
\patched{} & \textbf{100.0} & \textbf{100.0} & \textbf{100.0} & \textbf{100.0} && \textbf{47.2} & \textbf{92.6}  \\ \bottomrule
\end{tabular}
\caption{Applying patches on CheckLists. We see significant improvements when the patches apply and invariances when they do not apply or are unimportant. For Sentiment Analysis, the datasets are designed to evaluate patching with abstract conditions, thus we see no effects from using regex based patches. For testing invariance, we report the percentage of inputs for which the prediction did not change w.r.t.\ the base model.}
\label{tab:results-synth}
\end{table}

\section{Patching models on real benchmarks}
\label{sec:real_experiments}

\subsection{Sentiment Analysis}
Unless noted otherwise, all datasets in this subsection are derived from Yelp Review \citep{zhang2015}. To fix errors on low-accuracy slices, we write patches by inspecting a random subset of 10-20 errors made by \originalpf{}.

\paragraph{Controlling the model.} In order to check if patches can control model behavior with abstract conditions ``in the wild'', we manually annotate a random subset of 500 reviews with food and service specific sentiment (``The food was good, service not so much'' is labeled as {\cmss{service: 0, food: 1}}). We then construct override patches of the form \explanation{if \mybox{food / service} is good / bad, then label is positive / negative}, and evaluate models as to how often (on average) the prediction is as expected when the patch applies and how often it is unchanged when the patch does not apply. We present results in Table \ref{tab:results-steering}.
The sentiment of both aspects typically agrees, and thus even models without patching often behave according to the patch. We note that natural language patches improve patched behavior the most (when compared to baselines), while almost never changing predictions when the patch does not apply. We additionally present results only on the subset of our aspect annotated examples where both aspects \emph{disagree} in Table~\ref{tab:results-steering-add}. Overall, we see a more pronounced difference \iee{} our model gets a \textasciitilde{}27 point boost in accuracy when the patch condition applies, while maintaining invariance when the condition does not apply.

\begin{table}[]
\tiny
\ra{1.3}
\centering
\begin{tabular}{@{}lrr@{}} \toprule
Model       & Correctly patched (applies)& Invariance (does not apply) \\ \midrule
\original{}    & 91.5                                & n/a                                    \\
\originalpf{} & 91.1                                 & n/a                                    \\
\rbpatches{}  & 91.0                                 & 99.5                                   \\
\originalprompts{}     & 92.3                                 & 98.4                                   \\
\patched{}     & \textbf{95.8}                                 & 99.4    \\ \bottomrule     
\caption{To measure how well patches control behavior ``in the wild'', we evaluate the model's ability to match the label specified by the patch when it applies, and invariance w.r.t the base model when the patch does not apply, on a subset of yelp with sentiment annotations for different aspects}
\label{tab:results-steering}
\end{tabular}
\end{table}

\begin{table}[]
\scriptsize
\centering
\ra{1.3}
\begin{tabular}{@{}lrr@{}} \toprule
Model       & Correctly patched (applies)& Invariance (does not apply) \\ \midrule
\original{}    & 52.1                                & n/a                                    \\
\originalprompts{}     & 55.7                                 & 97.6                                   \\
\originalpf{} & 53.5                                 & n/a                                    \\
\rbpatches{}  & 55.1                                 & 100.0                                   \\
\patched{}     & \textbf{79.6}                                 & 99.4    \\ \bottomrule     
\caption{ We evaluate the model's ability to match the label specified by the patch when it applies, and invariance w.r.t the base model when the patch does not apply, on a subset of yelp with sentiment annotations for different aspects. In this table, we specifically consider inputs where both food and service aspects differ in sentiment.}
\label{tab:results-steering-add}
\end{tabular}
\end{table}

\paragraph{Patching low-accuracy slices.} We identify slices where our base model has (comparatively) low accuracy, and check whether patches can improve performance. 
{\dataset{Yelp-stars}} consists of all examples in Yelp Review with the word `star' present. For this subset, we use two overrides patch: \explanation{If review gives 1 or 2 stars, then label is negative}, \explanation{If review gives 0 stars, then label is negative}.
{\dataset{Yelp-Colloquial}} is a label-balanced slice consisting of examples having the colloquial terms \{dope, wtf, omg, the shit, bomb, suck\}. Because the colloquial use of these terms depends on context, we further construct {\dataset{Yelp-Colloquial-Control}}, a CheckList where the same terms are used in their traditional sense (\egg{} ``The manager was a dope'', ``The bomb was found by the police at the restaurant''). A model can do well on both of these datasets simultaneously only if it understands the contextual nuance associated with colloquial terms, rather than relying on simple shortcuts such as equating ``bomb'' with ``good''. For these datasets, we write simple feature-based patches such as \explanation{If food is described as bomb, then food is good} for each term.
Finally, we use the ``Women's E-commerce Clothing Reviews'' dataset ({\dataset{WCR}}) from \citet{zhongmeta21} and add two override patches: \explanation{If review mentions phrases like needs to be returned, then label is negative}, and \explanation{If fit is boxy, then label is negative}.

In Table~\ref{tab:results-realsentiment}, we observe that a very small number of language patches improve performance by 0.5-4.1 accuracy points, always outperforming both the original model and baselines. 
These gains are not a result of the added synthetic data, as \originalpf{} often \emph{lowers} performance.
Qualitatively, \originalprompts{} tends to rely on shortcuts such as copying over the label in the patch rather than gating and integrating the information, while \rbpatches{} cannot deal with simple semantic understanding, \egg{} the rule on {\dataset{Yelp-stars}} fires for \textit{``Will deduct 1 star for the service but otherwise everything was excellent''}, leading to an incorrect patch application.
Natural language patches avoid both of these pitfalls by explicitly modeling gating and feature interpretation with learnt models.

\begin{table}[]
\tiny
\ra{1.3}
\centering
\begin{tabular}{@{}lrrrr@{}} \toprule
Model         & \dataset{Yelp-Stars} & \dataset{Yelp-Colloquial} & \dataset{Yelp-Colloquial-Control} & \dataset{WCR} \\ \midrule

\original{}     & 93.1 & 89.1 & 100.0 & 89.6   \\
\originalpf{} & 93.6 & 88.6 & 100.0 & 88.9  \\
\rbpatches{} & 92.7 & 91.9 & 88.1 & 90.0   \\
\originalprompts{} & 90.8 & 85.2 & 70.1 & 88.3  \\
\patched{} & \textbf{94.5} & \textbf{93.2} & \textbf{100.0}  & 90.1  \\ \bottomrule

\end{tabular}
\caption{Using Override and Feature Based patches to fix bugs on various benchmarks derived from real sentiment analysis datasets. For {\dataset{Yelp-Colloquial}}, we also generate an control test based on CheckList.}
\label{tab:results-realsentiment}
\end{table}

\subsection{Spouse Relation Extraction}

We construct {\spousetest{}}, an out-of-distribution test benchmark derived from FewRel \cite{gao-etal-2019-fewrel} by sampling from all relation types where at least one of the entities is a person ($n=8400$), and labeling examples as positive if they have the \texttt{Spouse} relation, negative otherwise.
We inspect 20 randomly sampled errors made by \originalpf{} on {\spousetest}, and observe that the model often confuses ``Entity1 has a child with Entity2'' with ''Entity1 is the child of Entity2'', and also misclassifies widowhood as negative. Thus, we write override patches for both of these error categories, resulting in 7 patches, presented in Table~\ref{tab:results-spouse-real}. Using all patches, we observe a \textasciitilde{}7.4 point F1 improvement over \original{}, while baselines either decrease F1 or barely improve it.

We highlight in Table \ref{tab:results-spouse-real} a phenomenon where each natural language patch in \emph{isolation} decreases performance, while all patches together increase performance. Further analysis reveals that this is because the gating head is not well calibrated in this case, and thus individual patches are applied incorrectly. However, the \emph{comparative} values of $g(x, c_i)$ are often ordered correctly, and thus a better patch is the one applied ($lp^{*}$ in Eq~\ref{eq:scalability}) when all patches are available.
We do further analysis in Table \ref{tab:results-spouse-real-gating-acc}, where we report the \emph{gating accuracy} (\iee{} whether the patch actually applies or not, labeled manually) of $lp^{*}$ on the subset of inputs where the \patched{} model changes the prediction (\texttt{Diff}), and where it changes the prediction to the correct label (\texttt{Diff} $\cap$ \texttt{Correct}).
With the caveat that patches are applied softly (and thus perfect gating accuracy is not strictly necessary), we observe that a few patches seem to hurt performance even in combination with others (\egg{} the first one). We also note that the patched model is right \emph{``for the right reasons''} in over 72\% of inputs where it changes the prediction to the correct one.

\begin{table}
\scriptsize
\centering
\ra{1.3}
\begin{tabular}{@{}llr@{}} \toprule
& Model & F1 \\ \midrule
& \original & 65.5 \\ 
& \originalpf & 61.4 \\   
& \rbpatches{} & 61.0 \\
& \originalprompts & 65.7 \\ 
& \patched{} & \textbf{72.9} \\ \midrule
\multirow{7}{*}{\rotatebox[origin=c]{90}{\textit{Using single patch}}} & \nexample{If $p_2$ is the son of $p_1$, then label is negative}     & 57.5   \\
& \nexample{If $p_1$ is the son of $p_2$, then label is negative}    & 58.9   \\
& \nexample{If $p_1$ and $p_2$ have a daughter, then label is positive}    & 61.9   \\
& \nexample{If $p_1$ and $p_2$ have a son, then label is positive}    & 66.8   \\
& \nexample{If $p_1$ is the widow of $p_2$, then label is positive}    & 
63.6   \\
& \nexample{If $p_1$ is the daughter of $p_2$, then label is negative}    & 50.7   \\
& \nexample{If $p_2$ is the daughter of $p_1$, then label is negative}    & 49.4 \\\bottomrule
\end{tabular}
\caption{Using Override Patches on {\spousetest} for Spouse relation extraction.}
\label{tab:results-spouse-real}
\end{table}

\begin{table}
\tiny
\centering
\ra{1.3}
\begin{tabular}{@{}lrr@{}} \toprule
Patch Condition & \texttt{Diff} & \texttt{Diff} $\cap$ \texttt{Correct}  \\ \midrule 
\nexample{$p_2$ is the son of $p_1$}     & 0.0 & NaN (0/0)  \\
\nexample{$p_1$ is the son of $p_2$}    & 75.0 & 75.0 \\
\nexample{$p_1$ and $p_2$ have a daughter} &  63.3 & 93.8   \\
\nexample{$p_1$ and $p_2$ have a son}  & 78.1 & 98.3  \\
\nexample{$p_1$ is the widow of $p_2$}    & 10.9 & 19.6 \\
\nexample{$p_1$ is the daughter of $p_2$}    & 71.4 & 100.0 \\
\nexample{$p_2$ is the daughter of $p_1$}    & 6.3 & 100.0 \\\midrule
Overall & 42.9 & 72.3 \\ \bottomrule     
\end{tabular}
\caption{We measure how often the chosen patch correctly applies to an input (\iee{} gating accuracy) for {\spousetest{}}, for the set of inputs where the patched model and original model differ (\texttt{Diff}) as well as the subset where the patched model is correct (\texttt{Diff} $\cap$ \texttt{Correct}).}
\label{tab:results-spouse-real-gating-acc}
\end{table}

\begin{figure*}
    \centering
    \begin{subfigure}{.245\textwidth}
  \centering
  \includegraphics[width=.9\linewidth]{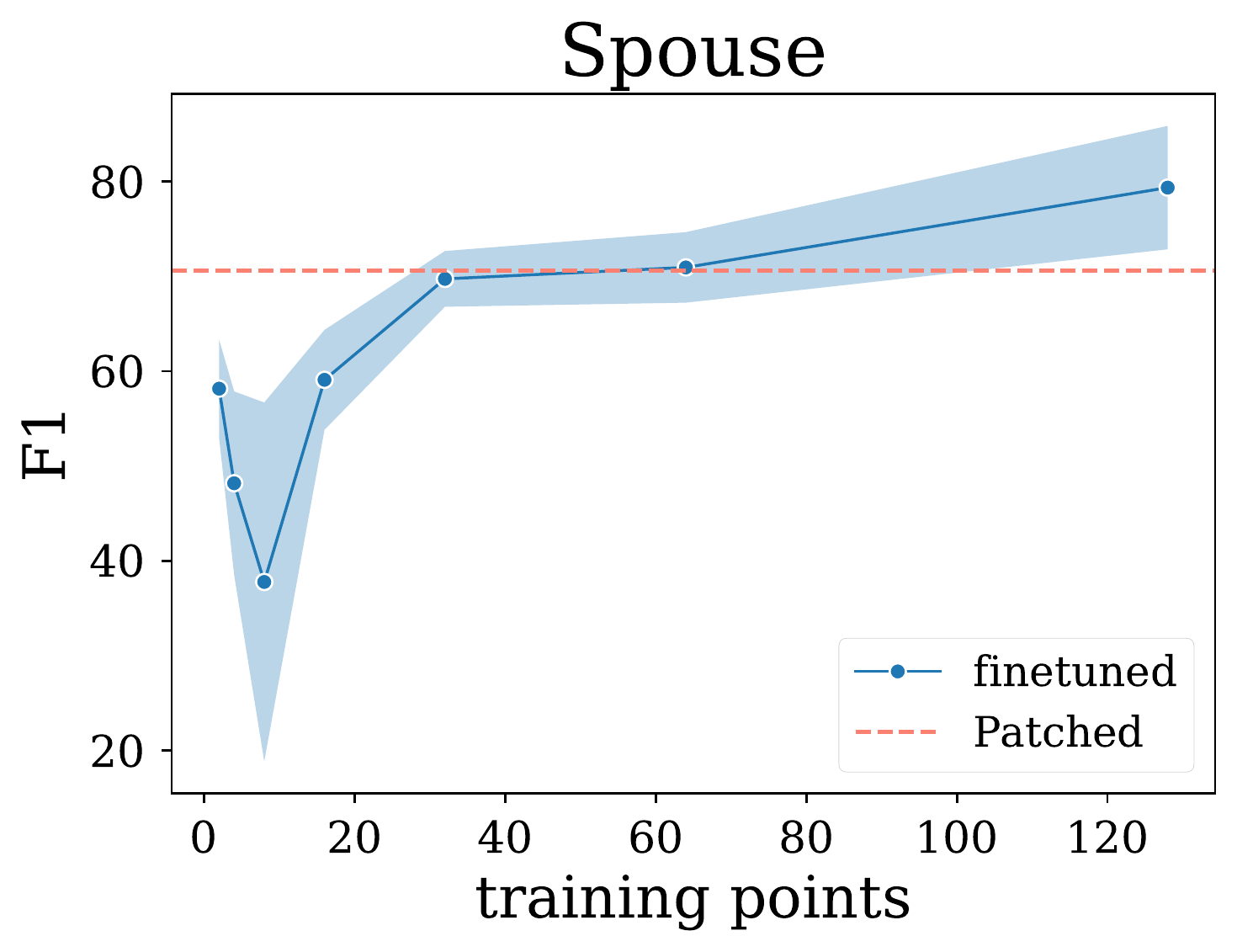}
  \label{fig:sfig1}
\end{subfigure}%
\begin{subfigure}{.245\textwidth}
  \centering
  \includegraphics[width=.9\linewidth]{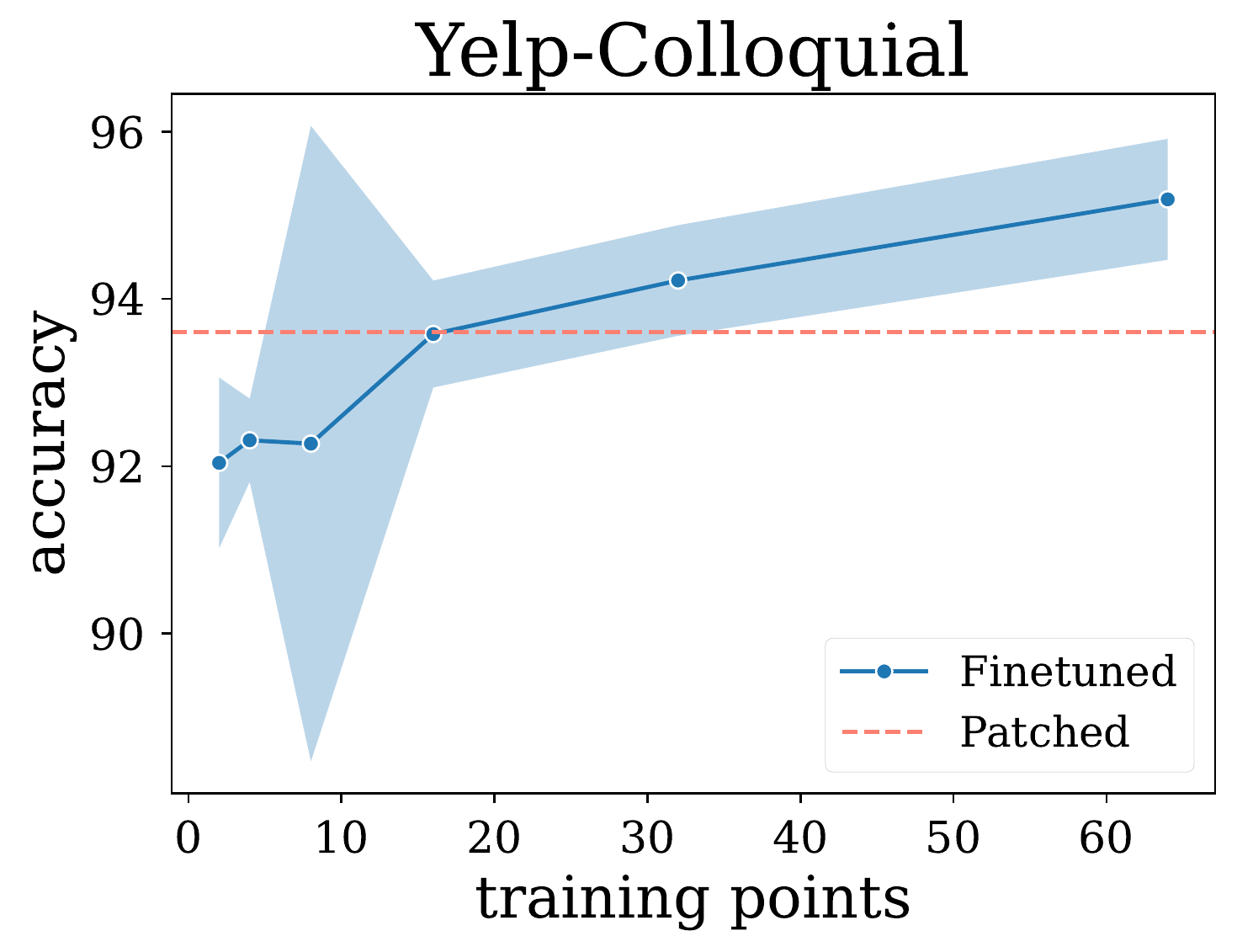}
  \label{fig:sfig2}
 \end{subfigure}
\begin{subfigure}{.245\textwidth}
  \centering
  \includegraphics[width=.9\linewidth]{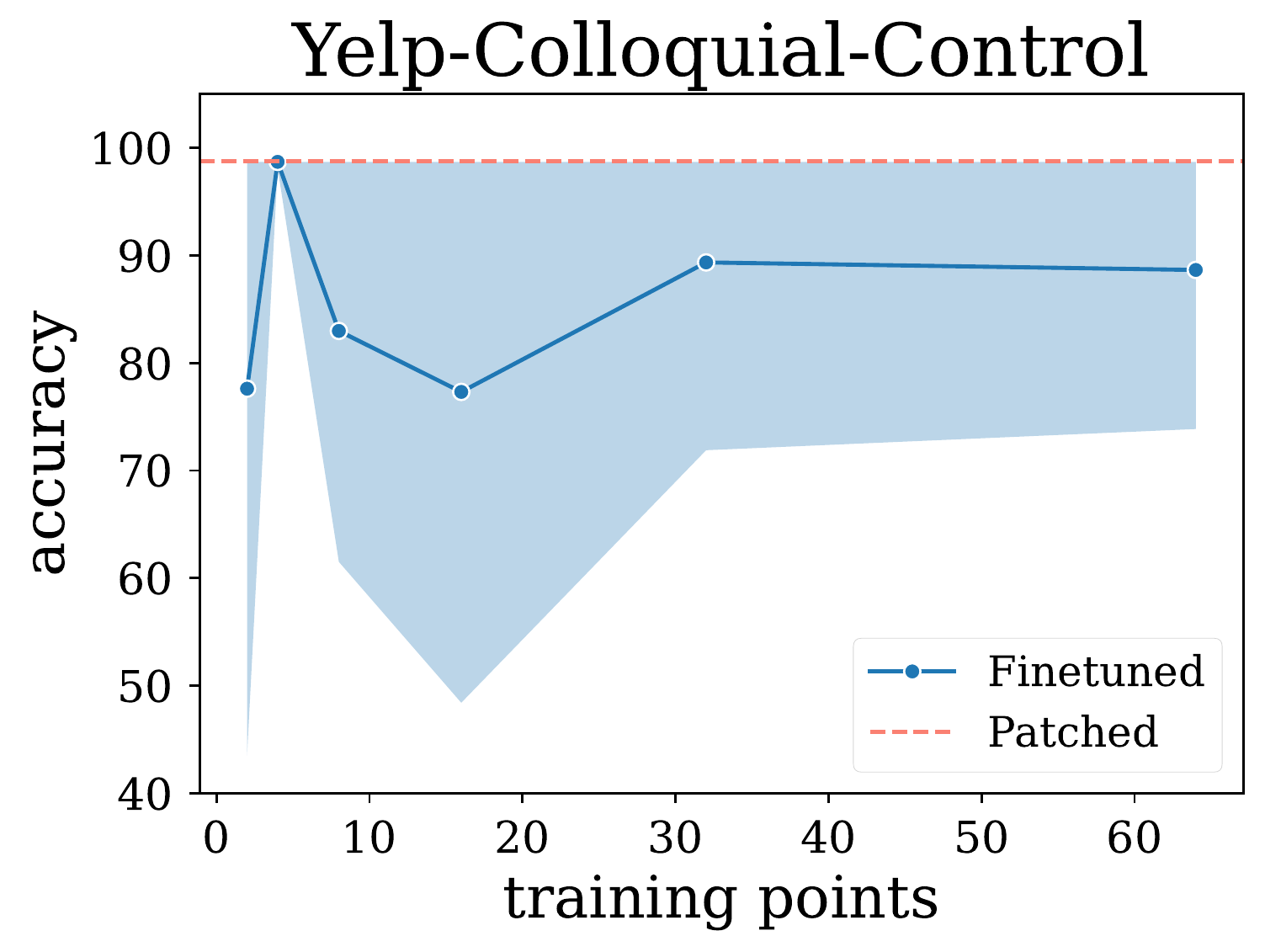}
  \label{fig:sfig3}
\end{subfigure}
    \begin{subfigure}{.245\textwidth}
  \centering
  \includegraphics[width=.9\linewidth]{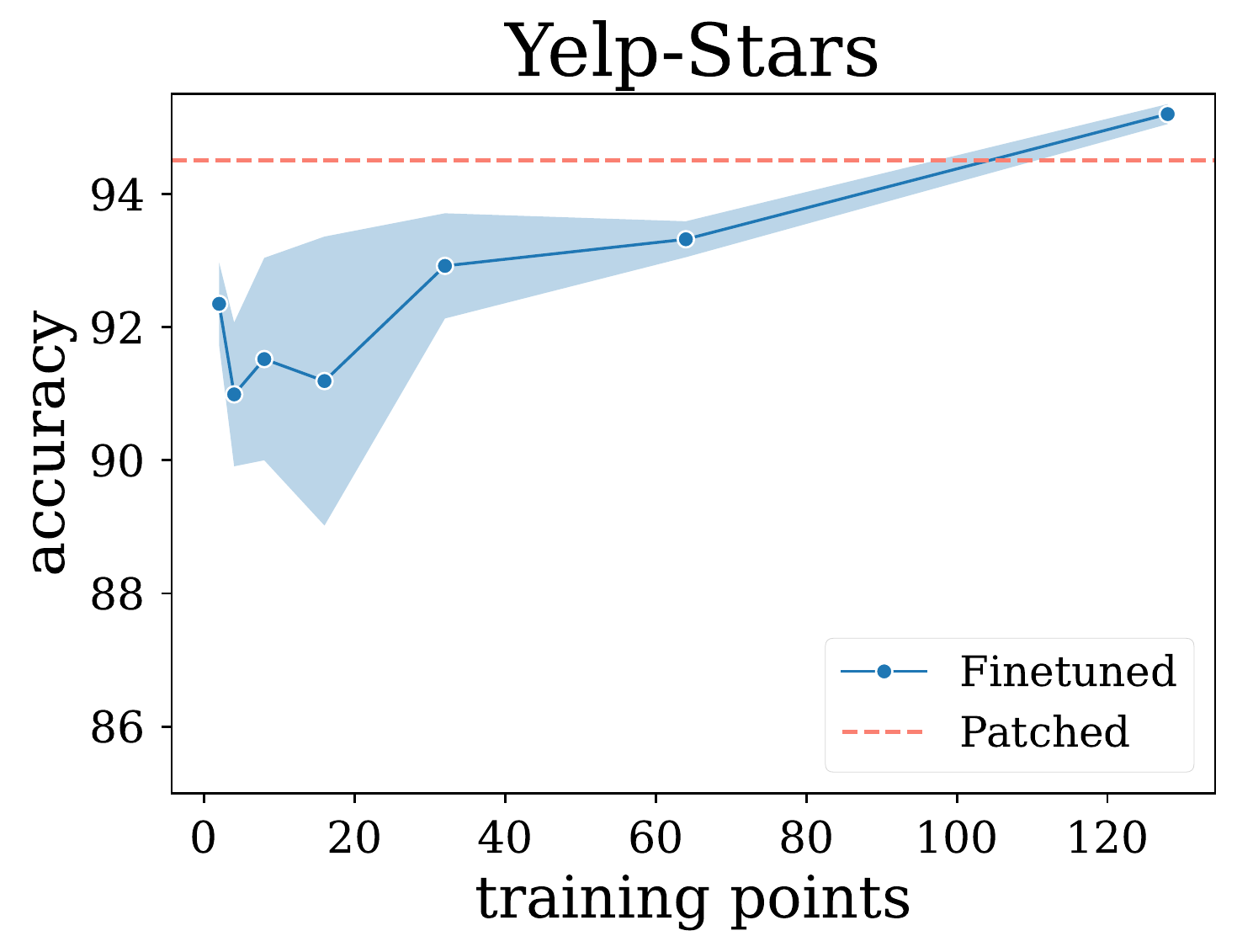}
  \label{fig:sfig4}
\end{subfigure}
    \caption{How many additional finetuning training examples it takes to reach the same accuracy level as patching. We report the mean and standard deviations across 5 runs.}
    \label{fig:ftvspatching}
\end{figure*}

\section{Analysis}
\subsection{How Important are EITs?}
\label{sec:eit_ablation}

\begin{table}
\scriptsize
\centering
\ra{1.3}
\begin{tabular}{@{}lrr@{}} \toprule
Patch Consequent                               & Patched  &  Patched (Without EITs) \\ \midrule
\nexample{$p_1$ went on a honeymoon with $p_2$}        & \textbf{59.1}  & 33.7 \\
\nexample{$p_1$ has kids with $p_2$}   & \textbf{75.2}  & 74.4  \\
\nexample{$p_1$ is engaged to $p_2$}   & \textbf{77.7}  & 64.8  \\        
\nexample{food is good}   & \textbf{67.8}  & 54.2  \\ 
\nexample{food is bad}   & \textbf{88.7}  & 56.5  \\ 
\nexample{service is good}   & \textbf{62.8}  & 52.9  \\  
\nexample{service is bad}   & \textbf{62.8}  & 52.9  \\ \midrule
Overall & \textbf{70.6} & 55.6
\\ \bottomrule
\end{tabular}
\caption{Patching Accuracy of a model with and without Entropy Increasing Transformations (EITs).}
\label{tab:results-eit}
\end{table}
The goal of Entropy Increasing Transformations (EITs; Section~\ref{sec:our_approach2}) is to prevent the interpreter head from learning shortcuts that either ignore patch features or rely exclusively on them.
We perform an ablation, comparing our model to a model trained without EITs on the CheckLists in Table \ref{tab:results-synth}  (Section~\ref{sec:synth_experiments}), where the feature-based patch consequent supplies important information for making a correct prediction.
From Table \ref{tab:results-eit}, we note that the interpreter head trained without EITs has much lower performance on these  datasets (as expected).

\subsection{Comparison to fine-tuning}
\label{sec:comparison_with_finetuning}

While patching is computationally lightweight, it requires domain knowledge or error analysis of incorrectly labeled examples. However, once such analysis is performed, one can label these additional examples and finetune the model on them. Ignoring the computational and infrastructure costs of repeated finetuning, for patching to be a competitive alternative to finetuning from an \emph{annotation budget} perspective, we require the gains from patching to only be matched by multiple labeled examples. To compare language patches with finetuning, we consider {\dataset{Yelp-stars}}, {\dataset{Yelp-Colloquial}}, and {\spousetest{}} and split each dataset into a training set with 128 examples, and a test set with remaining examples. Next, we finetune \original{}, on $k = \{2, 4, 8, 16, 32, 64, 128\}$ examples from the training set, stopping early if finetuning performance exceeds patched performance. We finetune for 64 steps and optimize using AdamW with a fixed learning rate of 1e-4. We report means and standard deviations obtained from finetuning with 5 random seeds. 

Results are presented in \figref{fig:ftvspatching}, where we note that over $100$ labeled examples are needed to match the performance of a single patch on {\dataset{Yelp-Stars}} or 7 patches on {\spousetest}.
On {\dataset{Yelp-Colloquial}}, the patched performance is matched with a mere $16$ examples. However, as noted earlier, {\dataset{Yelp-Colloquial}} is susceptible to simple shortcuts, and we observe that the performance on the control set {\dataset{Yelp-Colloquial-Control}} suffers significantly as we finetune on more data (with very high variance).
Thus, we conclude that language patches on these datasets are not only very efficient in terms of annotation effort (when compared to labeling data for finetuning), but also less susceptible to simple shortcuts that do not address the problem at the right level of abstraction.

\section{Related Work}
\paragraph{Learning with Language.} Natural language instructions or explanations have been used for training fewshot image classifiers \citep{Mu2020, Andreas2018}, text classifiers \cite{Zaidan2008, Srivastava2018, Camburu2018, Hancock2018, Murty2020}, and in the context of RL \citep{Branavan2012, goyal19, coreyes2019, mu2022improving}. All of these works are concerned with reducing labeled data requirements with language supervision, while our setting involves using language as a \emph{corrective} tool to fix bugs \emph{at test time}.

\paragraph{Prompt Engineering.} An emerging technique for re-purposing language models for arbitrary downstream tasks involves engineering ``prompts''. Prompts are high level natural language descriptions of tasks that allow developers to express any task as language modeling \citep{brown2020language,gao2021making, zhongmeta21}. While we could try and directly use prompting to incorporate language patches, our experiments show that the models we consider fail to correctly utilize patches in the prompt (Section \ref{sec:experiments}). With increasing scale models may gain the ability to interpret patches zero-shot, but qualitative exploration of the largest available models at the time of writing \cite[e.g. GPT-3;][]{brown2020language} indicates they still suffer from the same problem. 
Using patches for corrective purposes requires an accurate interpretation model, as well as ignoring the patch when it is not applicable. We solve these challenges by learning a gating head and an interpretation head through carefully constructed synthetic data. 

\paragraph{Editing Factual Knowledge.} Test time editing of factual knowledge in models is considered by \citet{Talmor2020, decao2021editing, mitchell2021fast, Meng2022LocatingAE}. Instead of modifying factual knowledge, we show that \emph{free-form language} patches can be used to fix \emph{bugs} on real data, such as correctly interpreting the meaning of the word ``bomb'' in the context of food or predicting that divorced people are no longer married.

\section{Conclusion}
When faced with the task of fixing \emph{bugs} in trained models, developers often resort to brittle regex rules or finetuning, which requires curation and labeling of data, is computationally intensive, and susceptible to shortcuts.
This work proposes \emph{natural language patches} which are declarative statements of the form ``if $\cond$, then $\cons$'' that enable developers to control the model or supply additional information with conditions at the right level of abstraction.
We proposed an approach to patching that models the task of determining if a patch applies (gating) separately from the task of integrating the information (interpreting), and showed that this approach results in significant improvements on two tasks, even with very few patches.
Moreover, we show that patches are efficient (1-7 patches are equivalent or better than as many as $100$ finetuning examples), and more robust to potential shortcuts.
Our system is a first step in letting users \emph{correct} models through a \emph{single step} ``dialogue''. Avenues for future work include extending our approach to a back-and-forth dialogue between developers and models, modeling pragmatics, interpreting several patches at once, and automating patch finetuning.

\section{Acknowledgements}
SM was partly funded by a gift from Apple Inc. We are grateful to Jesse Mu, Mirac Suzgun, Pratyusha Sharma, Eric Mitchell, Ashwin Paranjape, Tongshuang Wu, Yilun Zhou and the anonymous reviewers for helpful comments. The authors would also like to thank members of the Stanford NLP group and the Adaptive Systems and Interaction group at MSR for feedback on early versions of this work.

\section{Reproducibility}
Code and model checkpoints are available at \url{https://github.com/MurtyShikhar/LanguagePatching}.

\section{Limitations}

\paragraph{Scaling to large patch libraries.} For our approach, inference time scales linearly with the size of the patch library. This is primarily because the gating head makes predictions on each patch in our patch library (Eq~\ref{eq:scalability}). Instead of running the gating head on each patch, one can trade off exactness for efficiency, by running the gating head on a much smaller candidate set identified using fast approximate nearest neighbors \citep{johnson2019billion} on sentence embeddings.

\paragraph{Scaling to more patch types.} The current approach requires writing patch templates \emph{beforehand} based on prior knowledge of the kinds of corrective feedback that developers might want to write in the future. Writing patch templates manually is fundamentally bottlenecked by human creativity and foresight. Morever, since humans are required to write templates, it makes scaling up to different patch types harder, since we expect generalization to completely new patch \emph{types} to be poor \egg{} generalizing to a patch that requires counting. Future work can explore automatic generation of synthetic patch templates \egg{} using pre-trained language models.

\paragraph{Interpreting multiple patches.} Finally, the approach we develop can only incorporate a single patch at a time, by selecting the most relevant patch from our patch library. This precludes the model from being able to combine features from multiple patches---\egg{} \textit{``caviar is a kind of food''} and \textit{``If caviar is described as overpowering, then caviar is spoiled''}.

\bibliography{anthology,custom}
\bibliographystyle{acl_natbib}
\newpage
\clearpage

\appendix
\section{More details on Patch Finetuning}

\subsection{Sentiment Analysis Data}
\label{sec:appendix_sent}

The templates used for constructing inputs are in Table~\ref{tab:synth_inp}. We programmatically find all patches for an input, to generate labels.

\subsection{Relation Extraction Data}
\label{sec:appendix_re}
\paragraph{Override Patches.} Patches and Input templates for constructing patch finetuning data can be found in Table~\ref{tab:synthspousemoreinfo}. 

\paragraph{Feature Based Patches}
For training the gating head, we use the same data as generated by Table~\ref{tab:synthspousemoreinfo}. For training the interpreter head, we use patches and input templates in Table~\ref{tab:synthspousemoreinfofb} to generate finetuning data.

\subsection{Additional Finetuning Details}

After the model is finetuned in the Task finetuning stage, we finetune it additionally with a learning rate of learning rate of 1e-4 and with a linear warmup scheduler which ramps up the learning rate from $0$ to 1e-4 over $100$ steps. The training batch size is $32$, and we clip gradients to have a max norm of 5. We early stop based on validation performance on a held out subset of the patch finetuning data.

\begin{table}[]
\centering
\scriptsize
\ra{1.3}
\begin{tabular}{@{}l@{}}
\toprule
\multicolumn{1}{c}{\textbf{Override Patches}} \\ \midrule
\template{If [\mybox{Entity1}/\mybox{Entity2}] is not a person, then label is negative} \\
 \template{If \mybox{Entity1} is the [child / parent] of \mybox{Entity2}, then label is negative} \\ 
 \template{If \mybox{Entity1} and \mybox{Entity2} have children, then label is positive} \\ 
\template{If \mybox{Entity1} and \mybox{Entity2} are divorced, then label is negative} \\ 
\template{If \mybox{Entity1} is engaged to \mybox{Entity2}, then label is positive} \\ 
\template{If \mybox{Entity1} and \mybox{Entity2} are siblings, then label is negative} \\ \addlinespace
\multicolumn{1}{c}{\textbf{Feature Based Patches}} \\ \midrule
\template{If \mybox{cond}, then \mybox{Entity1} is [married/not married] to \mybox{Entity2}} \\ 
\template{If \mybox{cond}, then \mybox{Entity1} is divorced from \mybox{Entity2}} \\ 
\template{If \mybox{cond}, then \mybox{Entity1} is engaged to \mybox{Entity2}} \\ 
\template{If \mybox{cond}, then \mybox{Entity1} is the sibling of  \mybox{Entity2}} \\ 
\template{If \mybox{cond}, then \mybox{Entity1} is dating \mybox{Entity2}} \\ 
\template{If \mybox{cond}, then \mybox{Entity1} is the parent of \mybox{Entity2}} \\ \bottomrule
\end{tabular}
\caption{Patch templates used for the Patch Finetuning stage for relation extraction. Each \mybox{Entity} is sampled from a small list of names, and \mybox{cond} is a set of conditions derived from keywords.}
\label{tab:synthspouse}
\end{table}

\section{Patches used for Yelp-Colloquial.} 
We used the following patches for fixing bugs on Yelp-Colloquial:
\begin{itemize}
    \item \explanation{If clothes are described as dope, then clothes are good.}
    \item \explanation{If food is described as the shit, then food is good.}
    \item \explanation{If service is described as bomb, then service is good.}
    \item \explanation{If restaurant is described as bomb, then restaurant is good.}
    \item \explanation{If food is described as bomb, then food is good.}
    \item \explanation{If something is described as wtf, then something is bad.}
    \item \explanation{If something is described as omg, then something is good.}
    \item \explanation{If food is described as shitty, then food is bad.}
\end{itemize}

\section{More examples of Entropy Increasing Transformations}

To perform Entropy Increasing Transformations (EITs) for relation extraction, we convert \mybox{rel} (see Table~\ref{tab:synthspousemoreinfofb} into nonce words $\egg{}$ ``Alice has a kid with John'' gets transformed into ``Alice has a wug with John'', for which we use a patch ``If Entity1 has a wug with Entity2, then Entity1 and Entity2 have kids

\section{Regex Based Patches.}
The exact functions we use for patching with regexes can be found in Listing~\ref{lst:rbpatch_override} and Listing~\ref{lst:rbpatch_fbpatches}.

\section{Data Statistics for all evaluation slices}

Statistics for all slices used for evaluation can be found in Table~\ref{tab:stats}.

\begin{table}[]
\centering
\scriptsize
\ra{1.3}
\begin{tabular}{lr} \toprule
Dataset                   & \#examples \\ \midrule
{\dataset{Yelp-Stars}}                & 3172       \\
{\dataset{Yelp-Colloquial}}          & 1784       \\
{\dataset{WCR}}              & 2919       \\
{\dataset{Yelp-Colloquial (Control)}} & 67         \\
{\dataset{Yelp-Aspect}} & 439 \\
{\dataset{Spouse-NYT}}                & 8400    \\ \bottomrule   
\end{tabular}
\caption{Dataset statistics for all the real data slices considered in this work.}
\label{tab:stats}
\end{table}

\begin{table*}[]
\centering
\scriptsize
\ra{1.3}
\begin{tabular}{@{}l@{}}
\toprule
\template{[\mybox{Entity1}] [\mybox{rel}] [\mybox{Entity2}]} \\
\template{[\mybox{Entity1}] [\mybox{rel}] [\mybox{Entity2}] and [\mybox{Entity1}] is (not) married to [\mybox{Entity2}]} \\

\template{[\mybox{Entity1}] who [\mybox{rel}] [\mybox{Entity2}], [\mybox{rel2}] [\mybox{Entity3}]}
 \\ \addlinespace \midrule
\template{rel = [have-kids, are-engaged, is-sibling, is-parent]} \\
\template{Entity = [Alice, Bob, Stephen, Mary]} \\

\bottomrule
\end{tabular}
\caption{Templates used for constructing inputs for patch finetuning stage in relation extraction analysis. Terms marked with '()' are optional. \mybox{rel} is a list of 4 relation types. For each relation type, we have a small list of 3 to 4 words. For instance \texttt{have-kids} = [`has a kid with', `has a son with', `has a daughter with']}
\label{tab:synthspousemoreinfofb}
\end{table*}

\begin{table*}[]
\centering
\scriptsize
\ra{1.3}
\begin{tabular}{@{}l@{}}
\toprule
\template{The [\mybox{aspect}] at the restaurant was (\mybox{modifier}) (not) [\mybox{adj}]} \\
\template{The [\mybox{aspect}] was (\mybox{modifier}) (not) [\mybox{adj}]} \\
\template{The restaurant [has/had] (\mybox{modifier}) [\mybox{adj}] [\mybox{aspect}]} \\ 
\template{The [\mybox{aspect1}] was (not) [\mybox{adj1}], the [\mybox{aspect2}] was (not) [\mybox{adj2}]} \\
\template{The [\mybox{aspect1}] was (not) [\mybox{adj1}], but the [\mybox{aspect2}] was (not) [\mybox{adj2}]} \\
\template{The [\mybox{aspect1}] was really (not) [\mybox{adj1}], even though the [\mybox{aspect2}] was (not) [\mybox{adj2}]} \\ \addlinespace \midrule
\template{aspect = [food, service, ambience]} \\
\template{modifier = [really, surprising, quite]} \\

\bottomrule
\end{tabular}
\caption{Templates used for constructing inputs for patch finetuning stage in sentiment analysis. Terms marked with '()' are optional. \mybox{adj} comes from a small set of 6 positive and 6 negative adjectives, as well as 6 nonce adjectives for EITs}
\label{tab:synth_inp}
\end{table*}

\begin{table*}
\centering
\scriptsize
\ra{1.3}
\begin{tabular}{@{}ll@{}}
\toprule
& \textbf{Examples} \\ \midrule
                  
\multirow{6}{*}{\textit{Patches}} & \eexample{$e_0$}{Entity1 divorced Entity2}        \\ 

& \eexample{$e_1$}{Entity1 has kids with Entity2}    \\ 
& \eexample{$e_2$}{Entity1 is the parent of Entity2}    \\ 
& \eexample{$e_3$}{Entity1 and Entity2 are engaged}    \\ 
& \eexample{$e_4$}{Entity1 and Entity2 are just friends or coworkers}    \\ 
& \eexample{$e_5$}{Entity1 or Entity2 is not human}    \\ \midrule

\multirow{4}{*}{\textit{Inputs}} & \makecell[ml]{\inp{\highlight{Entity1} and \highlight{Entity2} have a kid named Person3.}{\posb{$e_1$}, \negb{$e_2$}} \\ \inp{Entity1 and \highlight{Entity2} have a kid named \highlight{Person3}.}{\posb{$e_2$}, \negb{$e_1$}}}         \\ \addlinespace
& \makecell[ml]{\inp{\highlight{Entity1} proposed to \highlight{Entity2}. The event was witnessed by Entity1's best friend Person3.}{\posb{$e_3$}, \negb{$e_4$}} \\\inp{Entity1 proposed to Entity2. The event was witnessed by \highlight{Entity1}'s best friend \highlight{Person3}.}{\posb{$e_4$}, \negb{$e_0$}}}         \\ \addlinespace

& \makecell[ml]{\inp{\highlight{Entity1} has decided to divorce \highlight{Entity2}. They have a child named Person3.}{\posb{$e_0$}, \negb{$e_3$}} \\ \inp{\highlight{Entity1} has decided to divorce Entity2. They have a child named \highlight{Person3}.}{\posb{$e_2$}, \negb{$e_0$}}} \\ \addlinespace

& \makecell[ml]{\inp{\highlight{Entity1} works at \highlight{location}.}{\posb{$e_5$}, \negb{$e_0$}} } \\\bottomrule

\end{tabular}
\caption{Patches along with a subset of inputs used for the Patch Finetuning stage for the {\spouse} relation extraction task. For each input, we highlight the two entities and provide examples of some \posb{positive} and \negb{negative} patches. }
\label{tab:synthspousemoreinfo}
\end{table*}

\begin{listing*}
\begin{minted}[linenos=true, fontsize=\footnotesize]{python}
def star_rbpatch(model, inp):
    keywords = [' 0 star', ' 1 star', ' 2 star', 
               ' zero star', ' one star', ' two star']
    patches = [(keyword, 0) for keyword in keywords]
    return sentiment_override_rbpatch(model, inp, patches)

def clothing_reviews_rbpatch(model, inp):
    return sentiment_override_rbpatch(model, inp, [('boxy', 0), ('needs to be returned', 0)])

def spousenyt(model, inp):
    patch_list = [('Entity1 is the son of Entity2', 0),
                  ('Entity2 is the son of Entity1', 0),
                  ('Entity1 and Entity2 have a son', 1),
                  ('Entity1 and Entity2 have a daughter', 1),
                  ('Entity1 is the daughter of Entity2', 0),
                  ('Entity2 is the daughter of Entity1', 0),
                  ('Entity1 is the widow of Entity2', 1)]
    return re_override_rbpatch(model, inp, patch_list)

# for all override patches 
def sentiment_override_rbpatch(model, inp, patch_list):
    '''
        inp: "X" a review for which we want to predict sentiment
        patch_list: list of override patches converted into a form (cond, label) where cond is
        a string condition and label is the associated binary label
    '''
    for cond, label in patch_list:
        if cond in inp:
            return label
    return model(inp)

# for override patches for relation extraction
def re_override_rbpatch(model, inp, patch_list):
    '''
        inp: "X. Entity1: e1. Entity2: e2"
        patch_list: list of override patches converted into a form (cond, label) where cond is
        a string condition and label is the associated binary label
    '''
    text, ent_info = inp.split(' Entity1:')
    e1, e2 = ent_info.split('. Entity2:')
    e1 = e1.strip()
    e2 = e2.strip()
    for cond, label in patch_list:
        p = patch.replace('Entity1', e1).replace('Entity2', e2)
        p2 = patch.replace('Entity1', '').replace('Entity2', '')
        if p in inp:
            return pred
        elif p2 in inp:
            return pred
    return model(x)

\end{minted}
\caption{Rule based override patching for all our experiments}
\label{lst:rbpatch_override}
\end{listing*}

\begin{listing*}
\begin{minted}[linenos=true, fontsize=\footnotesize]{python}
# Regex based patching for using feature based patches on 
# controlled experiments for sentiment analysis
def sentiment_regex_based(model, inp, patch_list):
    '''
        inp: "X. Entity1: e1. Entity2: e2"
        patch_list: list of feature patches of the form 'if aspect is word, then aspect is good/ bad'
        as a tuple (aspect, word, sentiment)
    '''

    for aspect, word, sentiment in patch_list:
        if '{} is {}'.format(aspect, word) in inp:
            inp = inp.replace(word, sentiment)
            break
    return model(inp)

# Regex based patching for controlled experiments on relation extraction
def re_regex_based(model, inp, patch_list):
    '''
        inp: "X. Entity1: e1. Entity2: e2"
        patch_list: list of feature patches converted into a form (cond, cons) where cond
        and cons are both strings
    '''
    text, ent_info = inp.split(' Entity1:')
    e1, e2 = ent_info.split('. Entity2:')
    e1 = e1.strip()
    e2 = e2.strip()
    for cond, cons in patch:
        p = cond.replace('Entity1', e1).replace('Entity2', e2)
        if p in inp:
            cons_curr = cons.replace('Entity1', e1).replace('Entity2', e2)
            inp = '{}. {} Entity1: {}. Entity2: {}'.format(cons_curr, text, e1, e2)
            break
    return model(inp)
            
# Regex based patching on yelp colloquial
def yelp_col_regex_based(model, inp, patch_list):
    if 'wtf' in inp:
        inp = inp.replace('wtf', 'bad')
    elif 'omg' in inp:
        inp = inp.replace('omg', 'good')
    elif 'the shit' in inp:
        inp = inp.replace('the shit', 'good')
    elif 'bomb' in inp and 'food' in inp:
        inp = inp.replace('bomb', 'good')
    elif 'bomb' in inp and 'service' in inp:
        inp = inp.replace('bomb', 'good')
    elif 'bomb' in inp and 'restaurant' in inp:
        inp = inp.replace('bomb', 'good')
    elif 'dope' in inp and 'clothes' in inp:
        inp = inp.replace('dope', 'good')
    elif 'sucks' in inp:
        inp = inp.replace('sucks', 'bad')
    return model(inp)
\end{minted}
\caption{Regex based patching for using feature based patches for all experiments.}
\label{lst:rbpatch_fbpatches}
\end{listing*}

\end{document}

%% file: macros.tex
\newcommand{\rmodel}{\ensuremath{g}}
\newcommand{\imodel}{\ensuremath{\mathcal{I}}}

\newcommand{\lp}{\ensuremath{lp}}
\newcommand{\cond}{\ensuremath{c}}
\newcommand{\cons}{\ensuremath{q}}

\DeclareMathOperator*{\rulef}{Rule}

\newcommand{\lpp}[1]{\ensuremath{lp_{#1}}}
\newcommand{\patchlib}{\ensuremath{P}}

\newcommand{\negex}[1]{\ensuremath{\textsc{neg}(#1)}}
\newcommand{\condp}[1]{\ensuremath{c_{#1}}}

\newcommand{\fix}{{\cmtt{Fix}}}
\newcommand{\fixed}[1]{\fix{(\model, $x$, #1)}}

\newcommand{\model}{\ensuremath{f}}

\newcommand{\lmss}[1]{\fontfamily{lmss}\selectfont{#1}}
\newcommand{\cmtt}[1]{\fontfamily{cmtt}\selectfont{#1}}
\newcommand{\cmss}[1]{\fontfamily{cmss}\selectfont{#1}}

\definecolor{bluish}{rgb}{0.875, 0.875, 0.95}

\definecolor{cexample}{rgb}{0.23, 0.30, 
0.45}

\definecolor{inpcolor}{rgb}{0.97, 0.30, 0.80}

\newcommand{\explanation}[1]{{\textit{``#1''}\xspace}}

\newcommand{\inp}[1]{\cmss{#1}\xspace}

\newcommand{\gap}{\vspace{0.2mm}}

\newcommand{\eexample}[2]{\gap\hspace*{1mm}\textcolor{cexample}{\textbf{{#1}}: {\cmss{#2}}}\gap}

\newcommand{\nexample}[1]{{{\cmss{#1}}}}

\newcommand{\egg}{e.g.,}
\newcommand{\iee}{i.e.,}

\newcommand{\original}{\textsc{Orig}}
\newcommand{\originalpf}{\textsc{Orig+PF}}
\newcommand{\originalprompts}{\textsc{Prompt}}

\newcommand{\rbpatches}{\textsc{Regex}}
\newcommand{\patched}{\textsc{Patched}}

\newcommand{\figref}[1]{Fig.~\ref{#1}}

\newcommand{\dataset}[1]{\lmss{#1}}

\newcommand{\spouse}{\dataset{Spouse}}
\newcommand{\spousetest}{\dataset{Spouse-FewRel}}

\newcommand{\spousesynth}{\dataset{Feat}}
\newcommand{\spousesynthc}{\dataset{Feat-Inv}}

\newcommand{\overrideabs}{\dataset{Override}}
\newcommand{\overrideabsneg}{\dataset{O-Inv}}
\newcommand{\interpretabs}{\dataset{Feat}}

\newcommand{\interpretabsn}{\dataset{Feat-Inv}}